# Intelligent real-time MEMS sensor fusion and calibration

Dušan Nemec, Aleš Janota, Marián Hruboš, and Vojtech Šimák

*Abstract*—This paper discusses an innovative adaptive heterogeneous fusion algorithm based on estimation of the mean square error of all variables used in real time processing. The algorithm is designed for a fusion between derivative and absolute sensors and is explained by the fusion of the 3-axial gyroscope, 3-axial accelerometer and 3-axial magnetometer into attitude and heading estimation. Our algorithm has similar error performance in the steady state but much faster dynamic response compared to the fixed-gain fusion algorithm. In comparison with the extended Kalman filter the proposed algorithm converges faster and takes less computational time. On the other hand, Kalman filter has smaller mean square output error in a steady state but becomes unstable if the estimated state changes too rapidly. Additionally, the noisy fusion deviation can be used in the process of calibration. The paper proposes and explains a real-time calibration method based on machine learning working in the online mode during run-time. This allows compensation of sensor thermal drift right in the sensor's working environment without need of re-calibration in the laboratory.

*Index Terms*—calibration, inertial navigation, mean square error methods, sensor fusion.

## I. INTRODUCTION

INERTIAL SENSORS manufactured by the MEMS (Micro Electro-Mechanical Systems) technology are the core of modern low-cost AHRSs (Attitude and Heading Reference Systems). The purpose of these systems is to determine rotation of the measured object with respect to the horizontal plane and northern direction which is a crucial task in mobile robotics, aviation, automated car navigation and many others. These sensor systems use nonlinear discrete numerical integration of the measured angular velocity which is a typical example of velocity measurement when the sensor is measuring time derivative of desired variable. Main disadvantage of this method is great sensitivity of the output quality to the precision of the sensor measurements (especially sensor bias will cause increasing drift of the integrated result). First step of elimination of the integrated error is calibration of the sensor. A standard way of calibration measures raw output of the sensor as a response to stimulus with known amplitude. Relation between raw and real sensor outputs is formed into the transfer function (calibration curve) and its parameters are obtained from the measurements during calibration in offline mode. It is possible to calibrate by:

--One point (zero order transfer function - bias only).
--Two points (first order transfer function - bias and gain),
--Multiple points (calibration curve is a polyline or higher order curve).

In order to eliminate influence of the sensor random noise each calibration point has to be computed as an average of multiple measurements in the same conditions [1]. This requires special laboratory equipment which provides accurate and steady simulation of different sensor stimuli. Zhang et al. proposed a method of estimation of the calibration constants for the 3-axial inertial sensor (gyroscope, accelerometer) [2]. Gyroscope bias is determined directly in a steady state and accelerometer bias is computed after multiple steps when the acceleration sensor is oriented vertically along each of its axes one by one or the sensor has to be exposed to precisely known stimuli [3][4]. All these methods are working in the offline mode. Wang and Hao proposed a method utilizing an artificial neural network combined with the Kalman filter for estimation of nonlinear calibration parameters [5]. For online calibration it is necessary to detect steady state of the object, e.g. by lower vibrations [6].

When MEMS sensors are used, their calibration parameters tend to drift with temperature [5] [7]. Transfer function is therefore two-dimensional – one input corresponds to raw sensor data and the second input is sensor temperature. Most of commercially available integrated MEMS sensors incorporate a temperature sensor which allows usage of advanced temperature compensation techniques.

In order to compensate integrated error during run-time it is necessary to use a secondary absolute sensor and provide a data fusion. The secondary sensor may be much noisier and have slower response but its error has to be kept inside fixed bounds. A modification of the Kalman filter can be used as a The paper was submitted for review in February 22, 2016. This work was supported by the European Regional Development Fund and the Ministry of Education of the Slovak Republic, within the project ITMS 26220220089 "New methods of measurement of physical dynamic parameters and interactions of motor vehicles, traffic flow and road".

Authors are with Dept. of Control and Information Systems, Faculty of Electrical Engineering, University of Žilina, Univerzitná 8215/1, Žilina 01026, Slovakia. e-mail:
{dusan.nemec; ales.janota; marian.hrubos; vojtech.simak}@fel.uniza.sk

core of the sensor fusion algorithm [3][8][9][23], however it might be difficult to estimate parameters of the filter (covariance matrix, state model) for a standalone sensor system because the Kalman filter parameters depend on the measured system. Another sensor fusion utilizes Bayesian networks and the stochastic approach [10][11][12]. We have proposed a heterogeneous sensor fusion method for one differential sensor and one absolute sensor which requires only minimum count of parameters independently from the measured system. The performance of our algorithm will be compared with the performance of the extended Kalman filter (EKF) used in direct form described in [23].

Our real-time calibration method utilizes error estimate obtained as a side output from the sensor fusion algorithm. This approach eliminates the need of steady state detection and offline calibration. Since it can be running all the time when the sensor is in use our method should compensate long-term drifts continuously.

## II. Heterogeneous Fusion Algorithm Considering Quality

The method will be explained on the example of the fusion of the 3-axial gyroscope (velocity sensor), 3-axial accelerometer (absolute attitude sensor) and 3-axial magnetometer (absolute heading sensor). Sensor axes are orientated according to the NED convention (*x*-North or forward, *y*-East or right, *z*-Down), Euler angles are computed in the ZYX convention ($\alpha$ – Roll, $\beta$- Pitch, $\gamma$- Yaw). Attitude of object is then expressed by roll and pitch angles; heading is expressed by yaw angle.

In order to express the quality of estimation we will use the mean square error (MSE). In general the error model of the attitude estimation is nonlinear [13]. MSE of the directly measured data is considered constant and depends on the used sensor; MSE of a computed variable $y = f(x_1, x_2, …, x_N)$ is approximated by:

$$\text{MSE}(y) \approx \sum_{k=1}^{N} \left(\frac{\partial f}{\partial x_k}\right)^2 \text{MSE}(x_k). \quad (1)$$

### A. Estimating Euler angles from gyroscope readings

In the inertial navigation the object's Euler angles are primary computed from the angular velocity $\omega$ measured by the gyroscope. There are several methods of angular velocity integration into Euler angles; we usually use the matrix-based algorithm. Rotation is expressed as the 3D transformation matrix $R$ updated by the infinitesimal update matrix computed from each sample of the angular velocity [14]:

$$\mathbf{R}_u = \mathbf{R}_{update} \cdot \mathbf{R}, \quad \text{where} \quad \mathbf{R}_{update} = \begin{bmatrix} 1 & \omega_z \Delta t & -\omega_y \Delta t \\ -\omega_z \Delta t & 1 & \omega_x \Delta t \\ \omega_y \Delta t & -\omega_x \Delta t & 1 \end{bmatrix} \quad (2)$$

and $\omega_x$, $\omega_y$, $\omega_z$ are the Cartesian components of the angular velocity vector and $\Delta t$ is a sampling period of the gyroscope.

The error of the gyroscope can be considered the same for each axis and constant:

$$\text{MSE}(\omega_i \Delta t) = \text{MSE}(\delta_i) = E_{\text{gyro}} = const \quad (3)$$

MSEs of the updated uncompensated rotational matrix $\mathbf{R}_u$ are:

$$\text{MSE}(R_{u1,k}) = \text{MSE}(R_{1,k}) + R_{2,k}^2 E_{\text{gyro}} + R_{3,k}^2 E_{\text{gyro}} + \\ + \delta_y^2 \text{MSE}(R_{3,k}) + \delta_z^2 \text{MSE}(R_{2,k}), \quad (4.1)$$

$$\text{MSE}(R_{u2,k}) = \text{MSE}(R_{2,k}) + R_{3,k}^2 E_{\text{gyro}} + R_{1,k}^2 E_{\text{gyro}} + \\ + \delta_z^2 \text{MSE}(R_{1,k}) + \delta_x^2 \text{MSE}(R_{3,k}), \quad (4.2)$$

$$\text{MSE}(R_{u3,k}) = \text{MSE}(R_{3,k}) + R_{1,k}^2 E_{\text{gyro}} + R_{2,k}^2 E_{\text{gyro}} + \\ + \delta_x^2 \text{MSE}(R_{2,k}) + \delta_y^2 \text{MSE}(R_{1,k}), \quad (4.3)$$

where index $k = 1 \div 3$. As can be seen, errors of all matrix elements are increasing with new samples. Uncompensated Euler angles are then [15]:

$$\alpha_{\text{gyro}} = \text{atan2}(R_{u2,3}, R_{u3,3}), \quad (5.1)$$
$$\beta_{\text{gyro}} = \arcsin(-R_{u1,3}), \quad (5.2)$$
$$\gamma_{\text{gyro}} = \text{atan2}(R_{u1,2}, R_{u1,1}). \quad (5.3)$$

Corresponding MSEs of the Euler angles are:

$$\text{MSE}(\alpha_{\text{gyro}}) = \frac{R_{u3,3}^2 \text{MSE}(R_{u2,3}) + R_{u2,3}^2 \text{MSE}(R_{u3,3})}{\left(R_{u2,3}^2 + R_{u3,3}^2\right)^2}, \quad (6.1)$$

$$\text{MSE}(\beta_{\text{gyro}}) = \frac{\text{MSE}(R_{u1,3})}{1 - R_{u1,3}^2}, \quad (6.2)$$

$$\text{MSE}(\gamma_{\text{gyro}}) = \frac{R_{u1,1}^2 \text{MSE}(R_{u1,2}) + R_{u1,2}^2 \text{MSE}(R_{u1,1})}{\left(R_{u1,1}^2 + R_{u1,2}^2\right)^2}. \quad (6.3)$$

These formulas are undefined at the gimbal lock ($\cos \beta = 0$ and $R_{u1,3} = \pm 1$). If such condition occurs it is impossible to determine both roll and yaw (one has to be chosen) and MSE estimation is very imprecise.

### B. Estimating roll and pitch from accelerometer readings

Secondary, the attitude of the object can be obtained from acceleration readings by formulas [14]:

$$\alpha_{\text{acc}} = \text{atan2}(-a_y, -a_z), \quad (7)$$

$$\beta_{\text{acc}} = \text{atan2}(a_x, \sqrt{a_y^2 + a_z^2}), \quad (8)$$

where $a_x$, $a_y$, $a_z$ are the components of the acceleration measured by the accelerometer bound with a moving object. MSE of attitude estimation depends on the dynamics of the system (the vector $a$ measured by the accelerometer is a sum of the gravitational acceleration $g$ and the object's own acceleration including vibrations which might be useful in different applications [16]). If the object is steady, variance of the vector $a$ is smaller and attitude estimation is more precise:

$$\text{MSE}(\alpha_{\text{acc}}) = \frac{a_y^2 \text{MSE}(a_z) + a_z^2 \text{MSE}(a_y)}{\left(a_y^2 + a_z^2\right)^2}, \quad (9)$$

$$\text{MSE}(\beta_{\text{acc}}) = \\ = \frac{\left(a_y^2 + a_z^2\right)^2 \text{MSE}(a_x) + a_x^2 \left[a_y^2 \text{MSE}(a_y) + a_z^2 \text{MSE}(a_z)\right]}{\left(a_y^2 + a_z^2\right)\left(a_x^2 + a_y^2 + a_z^2\right)^2} \quad (10)$$

In order to decrease error of estimation the acceleration can be averaged from multiple samples (oversampled measurement). Then MSEs of average acceleration components are:

$$\text{MSE}(\bar{a}_i) = \frac{1}{N}\sum_{k=1}^{N}(a_i[k]-\bar{a}_i)^2 + \frac{\text{MSE}(a_i)}{N}$$
$$= \frac{1}{N}\sum_{k=1}^{N}a_i^2[k] - \left(\frac{1}{N}\sum_{k=1}^{N}a_i[k]\right)^2 + \frac{\text{MSE}(a_i)}{N}. \quad (11)$$

The second expression allows processing of accelerometer samples by batches with size $N$ with lower memory requirements. Accelerometer's own errors $\text{MSE}(a_i)$ are negligible at higher $N$ with respect to the errors caused by vibrations. However, the simulations have shown that using batches is causing relatively large step changes in the resultant estimated variable. Therefore it is better to use online approximation of average and MSE; then the fusion can be performed in each step [17]:

$$\bar{s}_i[n] \approx \frac{(N-1)\bar{s}_i[n-1] + a_i[n]^2}{N}, \quad (12.1)$$

$$\bar{a}_i[n] \approx \frac{(N-1)\bar{a}_i[n-1] + a_i[n]}{N}, \quad (12.2)$$

where $\bar{s}_i[n]$ is a mean square of the $i$-th acceleration component in the $n$-th step. Formulas (12) are first order low-pass IIR filters with cut-off frequency:

$$f_{\text{cut}} = \frac{f_{\text{sample}}}{2\pi}\arccos\left(1 - \frac{1}{2N(N-1)}\right). \quad (13)$$

MSE of the estimated average $\bar{a}_i[n]$ is then:

$$\text{MSE}(\bar{a}_i[n]) \approx \bar{s}_i[n] - \bar{a}_i[n]^2 + \frac{\text{MSE}(a_i[n])}{N}. \quad (14)$$

In order to compute yaw from the magnetic induction vector $\boldsymbol{B}$, it has to be rotated from objects' local coordinates to the global horizontal plane by following:

$$\begin{bmatrix}B'_x\\B'_y\\B'_z\end{bmatrix} = \begin{bmatrix}\cos\beta & \sin\alpha\sin\beta & \cos\alpha\sin\beta\\ 0 & \cos\alpha & -\sin\alpha\\ -\sin\beta & \sin\alpha\cos\beta & \cos\alpha\cos\beta\end{bmatrix}\cdot\begin{bmatrix}B_x\\B_y\\B_z\end{bmatrix}. \quad (15)$$

Yaw is then computed by the formula:
$$\gamma_{\text{mag}} = \text{atan2}(-B'_y, B'_x). \quad (16)$$

The vertical component of the magnetic induction $B'_z$ is not used; therefore the third row of the matrix in (15) can be omitted in the algorithm. Since the transformation (15) is using estimated roll and pitch, quality of the yaw estimation depends on the quality of the attitude estimation.

$$\text{MSE}(B'_x) = (\cos\beta)^2\text{MSE}(B_x) + (\sin\alpha\sin\beta)^2\text{MSE}(B_y) +$$
$$+ (\cos\alpha\sin\beta)^2\text{MSE}(B_z) +$$
$$+ (B_y\cos\alpha\sin\beta - B_z\sin\alpha\sin\beta)^2\text{MSE}(\alpha) +$$
$$+ (B_x\sin\beta - B_y\sin\alpha\cos\beta - B_z\cos\alpha\cos\beta)^2\text{MSE}(\beta) \quad (17.1)$$

$$\text{MSE}(B'_y) = (\cos\alpha)^2\text{MSE}(B_y) + (\sin\alpha)^2\text{MSE}(B_z) +$$
$$+ (B_z\cos\alpha + B_y\sin\alpha)^2\text{MSE}(\alpha). \quad (17.2)$$

Then MSE of the yaw estimated by the magnetometer is:

$$\text{MSE}(\gamma_{\text{mag}}) = \frac{B'^2_y\cdot\text{MSE}(B'_x) + B'^2_x\cdot\text{MSE}(B'_y)}{(B'^2_y + B'^2_x)^2}. \quad (18)$$

Error of the magnetometer is also the same for each axis and can be considered constant:
$$\text{MSE}(B_i) = E_{\text{mag}}. \quad (19)$$

### C. Heterogeneous fusion algorithm

The scheme shown in Fig. 1 describes the algorithm that can be used for fusion of the Euler angles computed from gyroscope, accelerometer and magnetometer readings.

The fusion algorithm is based on incremental compensation of difference between incrementally integrated Euler angles $\alpha_{\text{gyro}}$, $\beta_{\text{gyro}}$, $\gamma_{\text{gyro}}$ and absolute but noisy Euler angles $\alpha_{\text{acc}}$, $\beta_{\text{acc}}$, $\gamma_{\text{mag}}$. The adaptive gain block (see Fig. 1) optimizes the impact of each data source in order to increase resultant precision. A value with lower MSE has a higher effect to the result [18]. If the gain block has unit gain, the values obtained by gyroscope integration are not taken into account and result is equal to Euler angles obtained from the accelerometer and gyroscope.

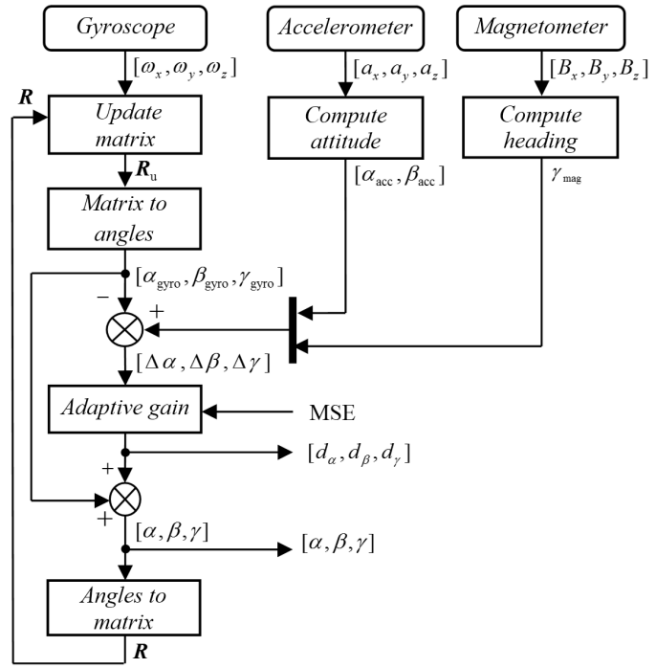

Fig. 1. The fusion algorithm scheme.

Resultant compensated Euler angles are equal to:

$$\begin{bmatrix}\alpha\\\beta\\\gamma\end{bmatrix} = \begin{bmatrix}\alpha_{\text{gyro}}\\\beta_{\text{gyro}}\\\gamma_{\text{gyro}}\end{bmatrix} + \begin{bmatrix}d_\alpha\\d_\beta\\d_\gamma\end{bmatrix} = \begin{bmatrix}K_\alpha\alpha_{\text{acc}} + (1-K_\alpha)\alpha_{\text{gyro}}\\K_\beta\beta_{\text{acc}} + (1-K_\beta)\beta_{\text{gyro}}\\K_\gamma\gamma_{\text{mag}} + (1-K_\gamma)\gamma_{\text{gyro}}\end{bmatrix}, \quad (20)$$

where $K_\alpha$, $K_\beta$, $K_\gamma$ are variable fusion gain coefficients adjusted according to the estimation errors and their values vary from 0 to 1. The vector $[d_\alpha, d_\beta, d_\gamma]$ represents fusion deviations:

$$\begin{bmatrix}d_\alpha\\d_\beta\\d_\gamma\end{bmatrix} = \begin{bmatrix}K_\alpha(\alpha_{\text{acc}} - \alpha_{\text{gyro}})\\K_\beta(\beta_{\text{acc}} - \beta_{\text{gyro}})\\K_\gamma(\gamma_{\text{mag}} - \gamma_{\text{gyro}})\end{bmatrix} \quad (21)$$

In case when fusion deviation $d_i$ is not available due to different sampling frequencies, it is simply considered zero. Usually the magnetometer has much lower sampling frequency than gyroscope or accelerometer. When the new sample from magnetometer is not available, the fusion deviation $d_\gamma$ is null.

Fusion deviations are important inputs for the automatic calibration algorithm described in the next chapter. Their errors are:

$$\text{MSE}(d_\alpha) = K_\alpha^2 \left[\text{MSE}(\alpha_{\text{acc}}) + \text{MSE}(\alpha_{\text{gyro}})\right], \quad (22.1)$$

$$\text{MSE}(d_\beta) = K_\beta^2 \left[\text{MSE}(\beta_{\text{acc}}) + \text{MSE}(\beta_{\text{gyro}})\right], \quad (22.2)$$

$$\text{MSE}(d_\gamma) = K_\gamma^2 \left[\text{MSE}(\gamma_{\text{mag}}) + \text{MSE}(\gamma_{\text{gyro}})\right]. \quad (22.3)$$

The fusion gain coefficients have to be adjusted in order to minimize the output error which is equal to:

$$\text{MSE}(\alpha) = K_\alpha^2 \cdot \text{MSE}(\alpha_{\text{acc}}) + (1-K_\alpha)^2 \cdot \text{MSE}(\alpha_{\text{gyro}}). \quad (23)$$

Note that this formula is also valid for pitch and yaw angles with corresponding coefficients. Output error is minimal when:

$$\frac{\partial \text{MSE}(\alpha)}{\partial K_\alpha} = 0 \ \wedge \ \frac{\partial^2 \text{MSE}(\alpha)}{\partial K_\alpha^2} > 0. \quad (24)$$

Solving these conditions we obtain the optimal gain:

$$K_\alpha = \frac{\text{MSE}(\alpha_{\text{gyro}})}{\text{MSE}(\alpha_{\text{gyro}}) + \text{MSE}(\alpha_{\text{acc}})}. \quad (25)$$

### D. Proof of Precision Improvement

We can substitute the optimal gain in the formula (23):

$$\text{MSE}(\alpha) = \frac{\text{MSE}(\alpha_{\text{gyro}}) \cdot \text{MSE}(\alpha_{\text{acc}})}{\text{MSE}(\alpha_{\text{gyro}}) + \text{MSE}(\alpha_{\text{acc}})}. \quad (26)$$

By dividing by $\text{MSE}(\alpha_{\text{gyro}})$ we can get:

$$\frac{\text{MSE}(\alpha)}{\text{MSE}(\alpha_{\text{gyro}})} = \frac{\text{MSE}(\alpha_{\text{acc}})}{\text{MSE}(\alpha_{\text{gyro}}) + \text{MSE}(\alpha_{\text{acc}})} < 1. \quad (27)$$

Since MSE is always higher than zero $\text{MSE}(\alpha)$ is always smaller than $\text{MSE}(\alpha_{\text{gyro}})$. Since the formula (26) is symmetrical, the same is valid for $\text{MSE}(\alpha_{\text{acc}})$. Therefore the theoretical fusion output error is always smaller than errors of any single estimation. Note that the output error directly depends on estimation of the MSEs during execution of the algorithm.

### E. Error of the Resultant Rotational Matrix

Fig. 1 contains the conversion block from compensated Euler angles to the rotational matrix [15]:

$$\boldsymbol{R} = \begin{bmatrix} c_\beta c_\gamma & c_\beta s_\gamma & -s_\beta \\ s_\alpha s_\beta c_\gamma - c_\alpha s_\gamma & s_\alpha s_\beta s_\gamma + c_\alpha c_\gamma & s_\alpha c_\beta \\ c_\alpha s_\beta c_\gamma + s_\alpha s_\gamma & c_\alpha s_\beta s_\gamma - s_\alpha c_\gamma & c_\alpha c_\beta \end{bmatrix}, \quad (28)$$

where $s_\alpha = \sin\alpha$, $c_\alpha = \cos\alpha$, etc. Note that the matrix $\boldsymbol{R} = \boldsymbol{R_u}$ when no magnetometer and accelerometer samples are available due to the different sampling frequency.

When a new sample is processed and errors of the compensated Euler angles are computed according to (26), it is possible to compute MSE of each element in the rotational matrix which is used in next sample processing (used in (4)):

$$\text{MSE}(R_{1,1}) = (\sin\beta\cos\gamma)^2 \text{MSE}(\beta) + (\cos\beta\sin\gamma)^2 \text{MSE}(\gamma), \quad (29.1)$$

$$\text{MSE}(R_{1,2}) = (\sin\beta\sin\gamma)^2 \text{MSE}(\beta) + (\cos\beta\cos\gamma)^2 \text{MSE}(\gamma), \quad (29.2)$$

$$\text{MSE}(R_{1,3}) = (\cos\beta)^2 \text{MSE}(\beta), \quad (29.3)$$

$$\text{MSE}(R_{2,1}) = (\sin\alpha\sin\gamma + \cos\alpha\sin\beta\cos\gamma)^2 \text{MSE}(\alpha) + \\ + (\sin\alpha\cos\beta\cos\gamma)^2 \text{MSE}(\beta) + \\ + (\cos\alpha\cos\gamma + \sin\alpha\sin\beta\sin\gamma)^2 \text{MSE}(\gamma), \quad (29.4)$$

$$\text{MSE}(R_{2,2}) = (\sin\alpha\cos\gamma - \cos\alpha\sin\beta\sin\gamma)^2 \text{MSE}(\alpha) + \\ + (\sin\alpha\cos\beta\sin\gamma)^2 \text{MSE}(\beta) + \\ + (\cos\alpha\sin\gamma - \sin\alpha\sin\beta\cos\gamma)^2 \text{MSE}(\gamma), \quad (29.5)$$

$$\text{MSE}(R_{2,3}) = (\cos\alpha\cos\beta)^2 \text{MSE}(\alpha) + (\sin\alpha\sin\beta)^2 \text{MSE}(\beta), \quad (29.6)$$

$$\text{MSE}(R_{3,1}) = (\cos\alpha\sin\gamma - \sin\alpha\sin\beta\cos\gamma)^2 \text{MSE}(\alpha) + \\ + (\cos\alpha\cos\beta\cos\gamma)^2 \text{MSE}(\beta) + \\ + (\sin\alpha\cos\gamma - \cos\alpha\sin\beta\sin\gamma)^2 \text{MSE}(\gamma), \quad (29.7)$$

$$\text{MSE}(R_{3,2}) = (\cos\alpha\cos\gamma + \sin\alpha\sin\beta\sin\gamma)^2 \text{MSE}(\alpha) + \\ + (\cos\alpha\cos\beta\sin\gamma)^2 \text{MSE}(\beta) + \\ + (\sin\alpha\sin\gamma + \cos\alpha\sin\beta\cos\gamma)^2 \text{MSE}(\gamma), \quad (29.8)$$

$$\text{MSE}(R_{3,3}) = (\sin\alpha\cos\beta)^2 \text{MSE}(\alpha) + (\cos\alpha\sin\beta)^2 \text{MSE}(\beta). \quad (29.9)$$

By comparing expressions in brackets with elements of the matrix $\boldsymbol{R}$ in (28) it is possible to simplify the formulas (29):

$$\text{MSE}(R_{1,1}) = (\sin\beta\cos\gamma)^2 \text{MSE}(\beta) + R_{1,2}^2 \text{MSE}(\gamma), \quad (30.1)$$

$$\text{MSE}(R_{1,2}) = (\sin\beta\sin\gamma)^2 \text{MSE}(\beta) + R_{1,1}^2 \text{MSE}(\gamma), \quad (30.2)$$

$$\text{MSE}(R_{1,3}) = (\cos\beta)^2 \text{MSE}(\beta), \quad (30.3)$$

$$\text{MSE}(R_{2,1}) = R_{3,1}^2 \text{MSE}(\alpha) + (\sin\alpha\cos\beta\cos\gamma)^2 \text{MSE}(\beta) + \\ + R_{2,2}^2 \text{MSE}(\gamma), \quad (30.4)$$

$$\text{MSE}(R_{2,2}) = R_{3,2}^2 \text{MSE}(\alpha) + (\sin\alpha\cos\beta\sin\gamma)^2 \text{MSE}(\beta) + \\ + R_{2,1}^2 \text{MSE}(\gamma), \quad (30.5)$$

$$\text{MSE}(R_{2,3}) = R_{3,3}^2 \text{MSE}(\alpha) + (\sin\alpha\sin\beta)^2 \text{MSE}(\beta), \quad (30.6)$$

$$\text{MSE}(R_{3,1}) = R_{2,1}^2 \text{MSE}(\alpha) + (\cos\alpha\cos\beta\cos\gamma)^2 \text{MSE}(\beta) + \\ + R_{3,2}^2 \text{MSE}(\gamma), \quad (30.7)$$

$$\text{MSE}(R_{3,2}) = R_{2,2}^2 \text{MSE}(\alpha) + (\cos\alpha\cos\beta\sin\gamma)^2 \text{MSE}(\beta) + \\ + R_{3,1}^2 \text{MSE}(\gamma), \quad (30.8)$$

$$\text{MSE}(R_{3,3}) = R_{2,3}^2 \text{MSE}(\alpha) + (\cos\alpha\sin\beta)^2 \text{MSE}(\beta). \quad (30.9)$$

Initial MSE of Euler angles should be initialized to a large value representing low quality, e.g.:

$$MSE(\alpha_0) = MSE(\gamma_0) = \pi^2 \quad MSE(\beta_0) = (\pi/2)^2. \quad (31)$$

In order to avoid extreme values of MSE (e.g. around gimbal lock singularity, see (6)), it is convenient to limit MSE of the rotational matrix to the interval (-1, 1). Maximal values for Euler angles' MSEs can be set according to (31).

### III. THE REAL-TIME CALIBRATION ALGORITHM

In this chapter we will discuss how fusion deviation (a side product of the fusion of two or more sensors) can be used for sensor calibration. It is convenient when error estimate of deviation is also available. The algorithm will be explained on the example of the gyroscope, accelerometer and magnetometer fusion proposed in the previous chapter.

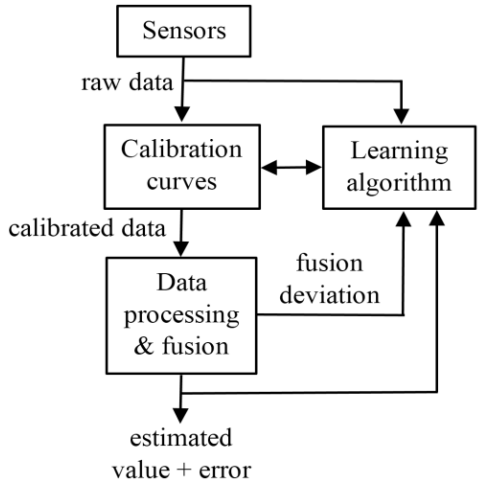

Fig. 2. The real-time calibration scheme in the context of the sensor data processing and fusion.

## A. Calibration model

A universal relation between the raw value $w$ measured by the MEMS sensor and the actual value $\omega$ of the measured variable is following:

$$\omega = C(w, T),  \quad (32)$$

where $T$ is sensor's temperature and $C(w,T)$ is a calibration transfer function [19]. Note that errors caused by the sensor's hysteresis are neglected [20].

Our algorithm assumes first order calibration transfer function at any given temperature, therefore there are two one-dimensional functions: gain $G(T)$ and bias $B(T)$ for which it is valid:

$$\omega = G(T) \cdot w - B(T) \quad (33)$$

Since the $C(w,T)$ function is continuous, partial functions $G(T)$ and $B(T)$ are also continuous and they can be expressed by polynomials:

$$G(T) = \sum_{k=0}^{N} g_k T^k, \quad (34)$$

$$B(T) = \sum_{k=0}^{N} b_k T^k. \quad (35)$$

where $g_k$ and $b_k$ are constants which change only by the long-term drift (e.g. by aging). In order to suppress sensitivity of the higher order coefficients, full scale of the sensor raw data and temperature should be normalized to the interval (0,1) or (-1,1). The proposed learning algorithm requires floating-point number implementation [21]. In order to maintain precision in full range and avoid ignoring small incremental steps (especially by 32-bit floating point IEEE754 format), the gain function should be shifted by one. Biases of the MEMS sensors drift faster than gain, therefore it is convenient to make the bias function independent from the gain function. The modified relation is:

$$\omega = [G(T) + 1] \cdot [w - B(T)]. \quad (36)$$

If the calibration formula (36) is used, the polynomial coefficients $g_k$ and $b_k$ can be initialized to zero.

In inertial navigation it is required to measure vectors of linear acceleration and angular velocity in all three axes (the 3-axial MEMS vibrational gyroscope and accelerometer are usually used for this purpose [4]). Each axis has its own gain and the offset calibration functions (independently from each other) according to (33). However, if the vector variable is measured, it is necessary to take sensor orientation into account. The most general case occurs when each axis is measured by a single physical sensor. The relation between the vector of measured raw data $\boldsymbol{w}$ and the calibrated vector $\boldsymbol{\omega}$ is following (in the orthonormal coordinate system):

$$\boldsymbol{\omega} = \boldsymbol{A} \cdot \boldsymbol{G}(T) \cdot [\boldsymbol{w} - \boldsymbol{B}(T)], \quad (37)$$

where $\boldsymbol{G}(T)$ is a diagonal matrix of the shifted gain functions and $\boldsymbol{B}(T)$ is a vector of the bias functions:

$$\boldsymbol{G}(T) = \begin{bmatrix} G_x(T)+1 & 0 & 0 \\ 0 & G_y(T)+1 & 0 \\ 0 & 0 & G_z(T)+1 \end{bmatrix}, \quad (38.1)$$

$$\boldsymbol{B}(T) = \begin{bmatrix} B_x(T) \\ B_y(T) \\ B_z(T) \end{bmatrix}. \quad (38.2)$$

Each of $G_i(T)$ and $B_i(T)$ functions are polynomials according to (34) and (35) respectively. For example, coefficients of the polynomial $G_x(T)$ are marked $g_{x,0}$ $g_{x,1}$ …, coefficients of the polynomial $B_x(T)$ are marked $b_{x,0}$ $b_{x,1}$ …etc.

The matrix $\boldsymbol{A}$ is a constant alignment matrix which considers misalignment between sensor's axes and the object's axes [22]. Since the gain matrix $\boldsymbol{G}(T)$ normalizes each axis separately, the alignment matrix has to be column-normalized:

$$\boldsymbol{A} = \begin{bmatrix} \sqrt{1 - a_{2,1}^2 - a_{3,1}^2} & a_{1,2} & a_{1,3} \\ a_{2,1} & \sqrt{1 - a_{1,2}^2 - a_{3,2}^2} & a_{2,3} \\ a_{3,1} & a_{3,2} & \sqrt{1 - a_{1,3}^2 - a_{2,3}^2} \end{bmatrix}. \quad (39)$$

## B. Backpropagation of the fusion error

Parameters $a_{i,k}$ $g_{i,k}$ $b_{i,k}$ have to be obtained by learning. Since the true value of the measured variable is unavailable during runtime and the data processing algorithm is nonlinear and recursive, we cannot minimize the error between raw readings of the sensor and the actual value as the static (offline) calibration does. However, it is possible to minimize the fusion deviation (see (21)) in long term. Fusion deviation is therefore used as an output error for learning. Implementation of the deterministic least-squares method for non-linear recursive system would require remembering all previous samples and would be very complicated. Such approach would be difficult to implement into low-cost hardware. Considering mentioned drawbacks, we have chosen to use a stochastic learning algorithm.

In order to decrease memory requirements and decrease the computational time for one iteration of the learning algorithm, we have decided to avoid recurrent learning. To be able to do that it is necessary to back-propagate the fusion deviation vector $[d_\alpha, d_\beta, d_\gamma]^T$ through the data processing algorithm to the angular velocity error vector $\boldsymbol{e} = [e_x, e_y, e_z]^T$. The data processing algorithm disregarding sensor fusion can also be modelled by the following continuous non-linear differential equation [15]:

$$\begin{bmatrix} \frac{d\alpha}{dt} \\ \frac{d\beta}{dt} \\ \frac{d\gamma}{dt} \end{bmatrix} = \begin{bmatrix} 1 & \frac{\sin\alpha \sin\beta}{\cos\beta} & \frac{\cos\alpha \sin\beta}{\cos\beta} \\ 0 & \cos\alpha & -\sin\alpha \\ 0 & \frac{\sin\alpha}{\cos\beta} & \frac{\cos\alpha}{\cos\beta} \end{bmatrix} \cdot \begin{bmatrix} \omega_x \\ \omega_y \\ \omega_z \end{bmatrix}, \quad (40)$$

The 3x3 matrix in (40) represents sensitivity of the output (Euler angles) to the calibrated gyroscope readings. The inverted matrix can be used to transform errors of the Euler angles (fusion deviations) to the errors of the calibrated angular velocity:

$$\begin{bmatrix} e_x \\ e_y \\ e_z \end{bmatrix} = \frac{1}{\Delta t_{cal}} \begin{bmatrix} 1 & 0 & -\sin\beta \\ 0 & \cos\alpha & \sin\alpha\cos\beta \\ 0 & -\sin\alpha & \cos\alpha\cos\beta \end{bmatrix} \cdot \begin{bmatrix} d_\alpha \\ d_\beta \\ d_\gamma \end{bmatrix}, \quad (41)$$

where $\Delta t_{cal}$ is a period of the calibration procedure (can be larger than the sampling period of the gyroscope). The relation (41) is valid only if the Euler angles $\alpha$, $\beta$, $\gamma$ are estimated precisely, which is achieved by compensation of the errors by sensor fusion. In order to avoid invalid adjustments of the calibration parameters caused by initial low quality of the estimated Euler angles, the calibration procedure is applied only if the fusion deviation is below some predefined threshold. This approach also eliminates the need for recurrent learning and the learning algorithm optimizes the error vector of the gyroscope (instead of direct optimization of the fusion deviation).

### C. Adaptive Scaling of the Fusion Deviation

Since the smaller calibration error will cause the smaller parameter change we can scale the fusion deviations according to their MSEs (22). The scaling function applied before backpropagation and learning can be e.g.:

$$d \leftarrow d \cdot \max\left(1 - \frac{\text{MSE}(d)}{e_{max}^2}, 0\right), \quad (42)$$

where $e_{max}$ is the maximal RMS of the fusion deviation acceptable for usage in calibration. This avoids degradation of calibration by initial low-quality data.

### D. The Learning Algorithm

It is possible to choose from many incremental learning algorithms. We have chosen the Adam algorithm [24], which is an extension of the well-known gradient descent algorithm. Its advantage over the standard gradient descent algorithm is its build-in 1st order infinite response filter for gradient and adaptive learning rate. As a result, the calibration parameters $c_{i,k}$ develop smoother than by the standard gradient descent method.
Goal of the learning is to minimize the loss function $E$:

$$E = \frac{1}{2}(\omega_{real} - \omega)^2 = \frac{1}{2}e^2. \quad (43)$$

The Adam algorithm requires gradients of the loss function by each parameter which is given by:

$$\frac{\partial E}{\partial c_{i,k}} = -e \cdot \frac{\partial \omega}{\partial c_{i,k}} \quad (44)$$

By differentiation of (37) we get:

$$\frac{\partial \omega}{\partial a_{1,2}} = (G_y(T) + 1)(w_y - B_y(T)) \cdot \begin{bmatrix} 1 & \frac{a_{1,2}}{a_{2,2}} & 0 \end{bmatrix}^T, \quad (45.1)$$

$$\frac{\partial \omega}{\partial a_{1,3}} = (G_z(T) + 1)(w_z - B_z(T)) \cdot \begin{bmatrix} 1 & 0 & \frac{a_{1,3}}{a_{3,3}} \end{bmatrix}^T, \quad (45.2)$$

$$\frac{\partial \omega}{\partial a_{2,1}} = (G_x(T) + 1)(w_x - B_x(T)) \cdot \begin{bmatrix} \frac{a_{2,1}}{a_{1,1}} & 1 & 0 \end{bmatrix}^T, \quad (45.3)$$

$$\frac{\partial \omega}{\partial a_{2,3}} = (G_z(T) + 1)(w_z - B_z(T)) \cdot \begin{bmatrix} 0 & 1 & \frac{a_{2,3}}{a_{3,3}} \end{bmatrix}^T, \quad (45.4)$$

$$\frac{\partial \omega}{\partial a_{3,1}} = (G_x(T) + 1)(w_x - B_x(T)) \cdot \begin{bmatrix} \frac{a_{3,1}}{a_{1,1}} & 0 & 1 \end{bmatrix}^T, \quad (45.5)$$

$$\frac{\partial \omega}{\partial a_{3,2}} = (G_y(T) + 1)(w_y - B_y(T)) \cdot \begin{bmatrix} 0 & \frac{a_{3,2}}{a_{2,2}} & 1 \end{bmatrix}^T, \quad (45.6)$$

$$\frac{\partial \omega}{\partial g_{i,k}} = (w_i - B_i(T)) \cdot T^k \cdot \begin{bmatrix} a_{1,i} & a_{2,i} & a_{3,i} \end{bmatrix}^T, \quad (46)$$

$$\frac{\partial \omega}{\partial b_{i,k}} = -(G_i(T) + 1) \cdot T^k \cdot \begin{bmatrix} a_{1,i} & a_{2,i} & a_{3,i} \end{bmatrix}^T. \quad (47)$$

where $G_i(T)$ is the value of the polynomial $G_i$ at the current normalized temperature $T$ etc. Speed of learning depends on the setting of learning rate. Because the bias of MEMS sensors drifts rapidly, learning rate of the coefficients $b_{i,k}$ should be highest. Since the alignment matrix is constant, learning rate of $a_{i,k}$ coefficients should be the smallest. This distribution avoids false changes in the alignment matrix according to the short-term drift of the sensor bias.

## IV. EXPERIMENTAL RESULTS

For evaluation of the proposed fusion and calibration algorithms we have used both the simulation model of the sensor and the real MEMS sensor. Simulation allows us to compare real bias and gain parameters with those obtained by learning. We have compared performance of the extended Kalman filter implemented according to [23], fusion with fixed gain (also known as the complementary filter) and our proposed fusion algorithm with adaptive gain based on estimation of MSE.

### A. Sensor Fusion Simulation

The first simulation analyzed a steady-state at nonzero attitude $\alpha = 30°$, $\beta = -45°$, $\gamma = 60°$. The simulated sensors' raw readings included noise according to Table 1; all sensors have used the output sample rate 512Hz. MSE of the sensor readings were considered to be equal to the square of RMS and time-invariant.

TABLE I
SIMULATED NOISE PARAMETERS

| Sensor | RMS | Bias |
|---|---|---|
| Gyroscope | 0.5 °/s | 20 °/s |
| Accelerometer | 1.0 m/s² | none |
| Magnetometer | 10 % of full range | none |

The value of the fixed fusion gain has been chosen according to the steady value of the adaptive gain (approx. 0.05 according to the Fig. 4). As can be seen in Fig. 3 the fusion algorithm with adaptive gain has shorter convergence time than the other methods. Faster convergence is caused by the peak of fusion gain (see Fig. 4) which is a result of initial low quality of the output (see (31)) and system increased fusion gain in order to compensate rapidly the initial errors by accelerometer. Results of fusion using fixed gain converge to the results of the fusion with adaptive gain after approx. 150 milliseconds. Output of the Kalman filter has much smoother response and lower error in a steady state.

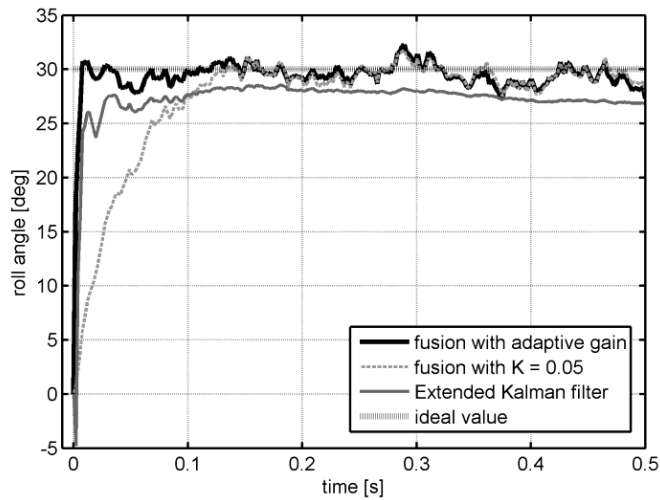

Fig. 3. The estimated roll angle in a steady state from noisy data with the fusion after reset.

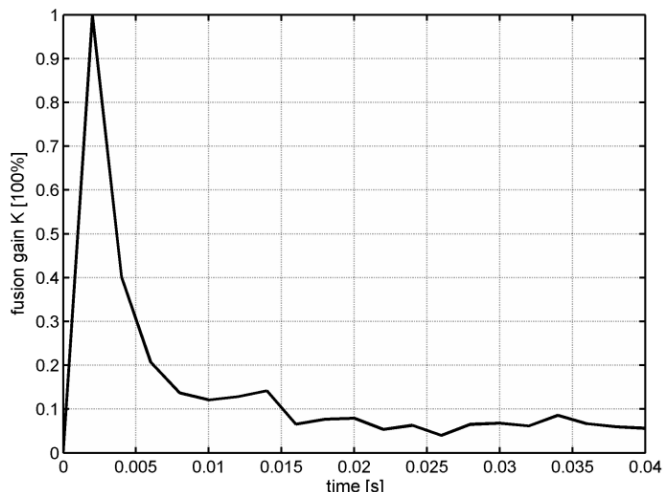

Fig. 4. The adaptive gain of the roll estimation fusion in a steady state after reset.

RMS errors of the estimated Euler angles in a steady attitude (100 seconds of experiment) are compared in Table 2. Execution time of the algorithm depends on implementation and hardware, therefore the values shown in Table 2 can be used only for relative comparison between discussed methods. All algorithms were implemented in the MATLAB environment; it is possible to decrease the execution time by using a compiled programming language like C and the code optimization.

TABLE II
COMPARISON OF FUSION ALGORITHMS IN A STEADY ATTITUDE

| Fusion algorithm | Roll RMS [deg] | Pitch RMS [deg] | Yaw RMS [deg] | Execution time [ms/sample] |
|---|---|---|---|---|
| Fixed gain | 1.17 | 1.06 | 2.31 | 0.21 |
| Adaptive gain | 1.09 | 0.93 | 1.56 | 0.23 |
| Kalman filter | 0.58 | 0.53 | 1.06 | 0.29 |

In order to verify dynamic parameters of the sensor fusion algorithm we have simulated harmonic rotation around axis $x$ (see Fig. 5). Due to the definition of Euler angles in Z-Y-X convention, the rotation around x-axis will affect only roll angle. Noise parameters of the sensors are the same as those used in the previous simulation. In order to visualize the difference between compared algorithms, Fig. 5 displays only the beginning of the experiment. Fig. 6 illustrates one important advantage of our proposed fusion algorithm over extended Kalman filter – stability in highly dynamic conditions. As can be seen the Kalman filter becomes instable after 6 periods of harmonic banking at frequency 1 Hz, but our proposed algorithm maintains its stability and precision. If we decrease the frequency of the banking (rotation) below 0.5 Hz, the instability of the extended Kalman filter disappears.

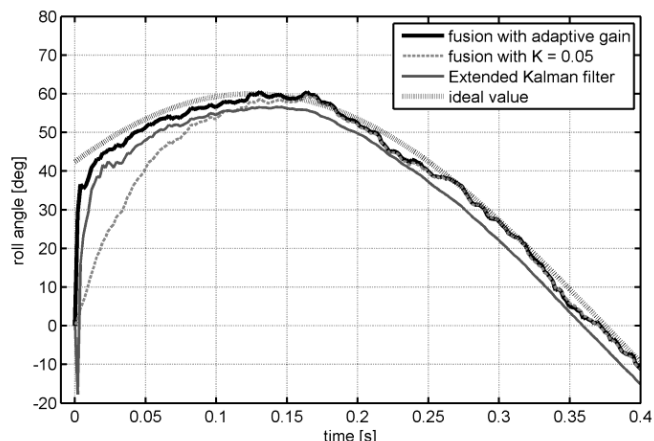

Fig. 5. The roll angle estimation during simulated periodic banking from -60° to +60°.

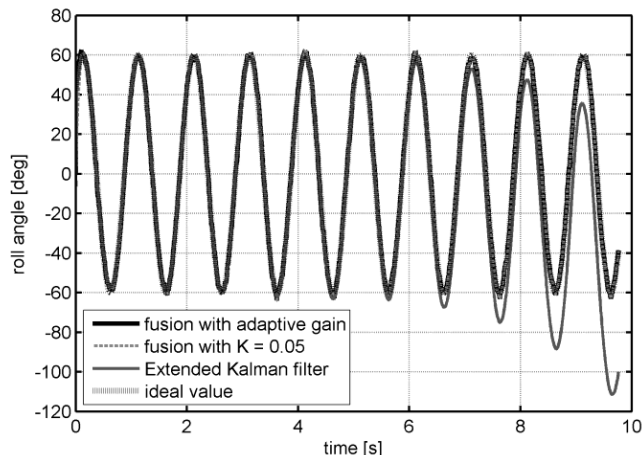

Fig. 6. Instability of the extended Kalman filter during periodic banking after longer period.

## B. Sensor Fusion Experiment

Experiments with the real MEMS sensor utilized a combined sensor board containing the 3-axial gyroscope and accelerometer IMU-3000 and the 3-axial magnetometer LSM303DLH). The noise parameters of the used sensor module are shown in Table 3.

TABLE III
REAL NOISE PARAMETERS

| Sensor | Full scale | RMS [x,y,z] |
|---|---|---|
| Gyroscope | ±500 °/s | [0.43, 0.41, 0.48] °/s |
| Accelerometer | ±8 g ≈ 78.5 m/s² | [0.65, 0.53, 0.51] m/s² |
| Magnetometer | 4 Gauss | [0.12, 0.19, 0.09] Gauss |

All noise parameters are estimated in a steady state. The bias of the accelerometer and magnetometer is considered to be zero (due to the previous static calibration).

The sensor board was banked by a servomotor from zero roll to approx. - 54 degrees (see Fig. 7). Filter parameter $N$ used in (12) and (14) was set to $N = 5$ which corresponds to the cut-off frequency $f_{cut} \approx 18$ Hz at the sampling rate 512 Hz. The absolute errors of the constant gain and the adaptive gain fusion are compared in Fig. 8. MSE of the Kalman filter is slightly higher (0,5%) than MSE of our proposed algorithm.

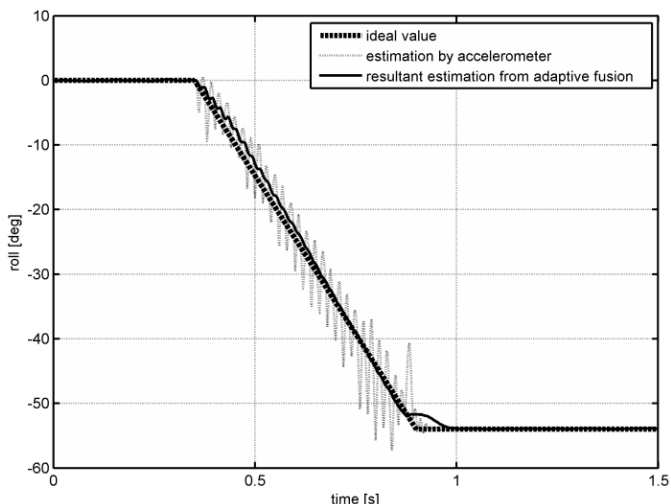

Fig. 7. Roll movement during the experiment compared with the values estimated from the accelerometer only and from the adaptive fusion.

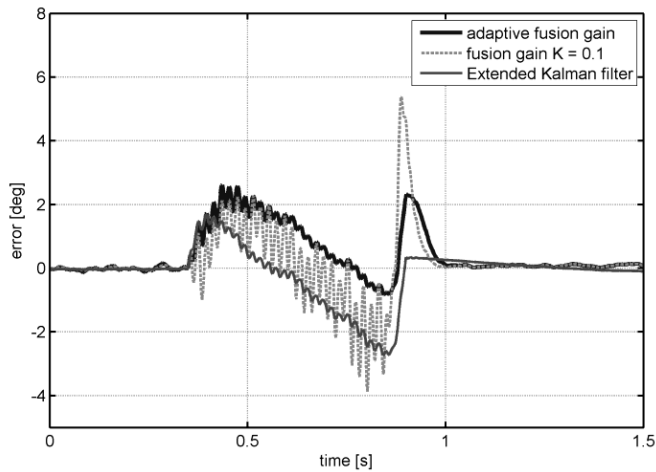

Fig. 8. Error of the estimated roll during the experiment.

Due to the vibrations caused by the servomotor during movement the adaptive fusion gain is lower while the servomotor is running (see Fig. 9). Note that the gain is stabilized in a steady state in the value $K = 0.1$ which was also used in the constant gain fusion in above comparison.

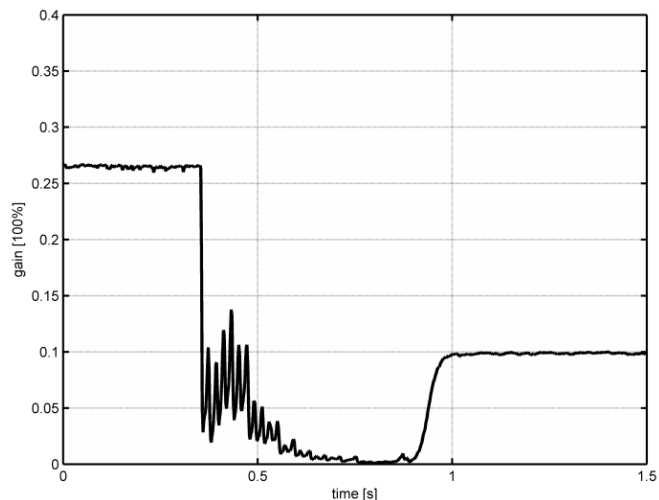

Fig. 9. The adaptive fusion gain during the experiment.

## C. Experimental Validation of the Calibration Algorithm

The proposed real-time calibration algorithm was used to estimate the bias and gain matrices of our gyroscope sensor module. Initial calibration parameters $g_{i,k}$ and $b_{i,k}$ were null. Since all sensors are placed on one board the sensor misalignment was neglected; therefore the $A$ matrix according to (39) was not needed to be adjusted. The thermal drifts of calibration parameters were not considered during the experiment because temperature of the sensor module was stable. If the general temperature-dependent version of the calibration algorithm is used the higher-order thermal coefficients $g_{i,k>0}$ and $b_{i,k>0}$ has to be adjusted slowly because their real values do not change rapidly during lifetime of the sensor. Mentioned higher-order calibration parameters allow faster adaptation to different temperature which can be useful in some applications (e.g. indoor-outdoor transition of a mobile robot).

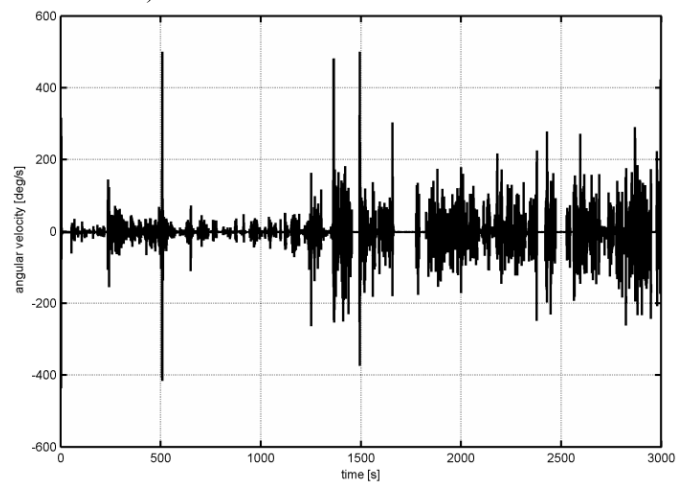

Fig. 10. Raw measured angular velocity around the x-axis during the experiment.

As an experimental input we have used a previously mentioned combined MEMS sensor module. The module was moved randomly with no pre-defined pattern (see Fig. 10 for an illustration) therefore the precise attitude and yaw are unknown. The raw (non-calibrated) accelerometer and magnetometer readings as a secondary input have been used in order to demonstrate robustness of the learning algorithm. The learning parameters were following:

TABLE IV
LEARNING PARAMETERS

| Learning Parameter | Symbol | Value |
| --- | --- | --- |
| Bias learning rate | $\lambda_{bias}$ | $10^{-6}$ |
| Gain learning rate | $\lambda_{gain}$ | $10^{-8}$ |
| 1st momentum filter [24] | $\beta_1$ | 0.999 |
| 2nd momentum filter [24] | $\beta_2$ | 0.9999 |
| Maximal fusion MSE | $e_{max}$ | 5 °/s |

The given learning rate is close to the upper limit for usage in real systems; since the whole fusion algorithm is iterative (closed loop) higher learning rates could cause instability. The lower learning rate will result in smoother but also slower bias development. Resultant bias development during automated learning is shown in Fig. 11.

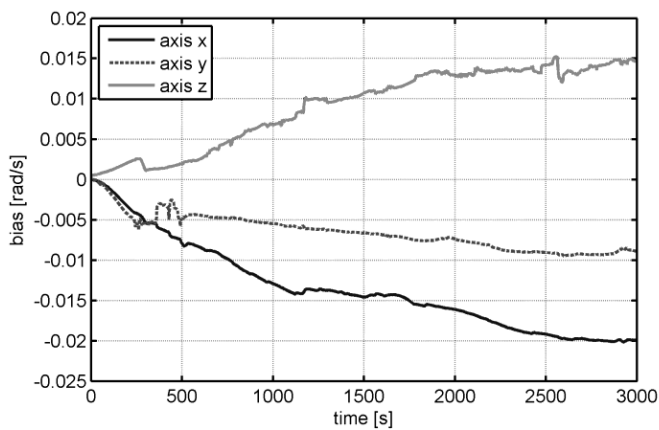

Fig. 11. Gyroscope bias real-time learning.

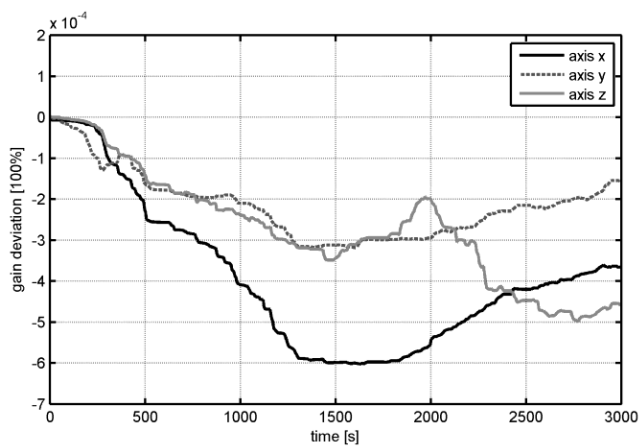

Fig. 12. Real-time learning of the gyroscope's gain.

Comparison with the actual bias values measured after the experiment by static calibration is in Table 5. As can be seen, the real-time calibration method converges to the real values.

The used gyroscope module has internally compensated gain therefore the gain deviance is very small (note scale of the y-axis in Fig. 12).

TABLE V
COMPARISON BETWEEN LEARNED BIAS AND VALUES ESTIMATED BY STATIC CALIBRATION

| Gyroscope axis | Learned bias [rad/s] | Bias obtained by static calibration [rad/s] | Error of the learned bias [% of the full scale] |
| --- | --- | --- | --- |
| $x$ | $-2.0 \cdot 10^{-2}$ | $-2.52 \cdot 10^{-2}$ | 0.05 % |
| $y$ | $-0.9 \cdot 10^{-2}$ | $-1.19 \cdot 10^{-2}$ | 0.03 % |
| $z$ | $1.5 \cdot 10^{-2}$ | $1.26 \cdot 10^{-2}$ | 0.03 % |

## V. DISCUSSION

As can be seen in Fig. 3 gyroscope bias is effectively suppressed by the sensor fusion. The fusion with the constant fusion gain causes smoother output but higher estimation error. The adaptive gain allows very fast reactions and start-up. According to Fig. 4 the adaptive algorithm reflects initial low quality of roll estimation; therefore the gain (weight of the attitude estimated by the accelerometer) is initially high and then rapidly decreases with the roll estimation error. If we compare the proposed algorithm with the widely-used extended Kalman filter, our algorithm converges faster and it is stable even during very dynamic changes. On the other hand, the extended Kalman filter has smoother response and lower RMS. Another advantage of our algorithm comes from its lower execution time.

Experiments with the real MEMS sensors approved results obtained by simulations. The MSE estimation algorithm is able to detect rapid changes in movement including vibrations (see Fig. 9) which adaptively decreases the influence of the accelerometer (absolute but noisy sensor) to the result. Estimation of MSE during algorithm execution also provides additional valuable information about the quality of the result. If additional information about the object's state is available, it is possible to change input MSE of the accelerometer to reflect known systematic errors.

Second experiment series evaluated the automated calibration algorithm. According to Fig. 11 the bias learning works also during movement and slowly converges to the precise bias values. Disadvantage of the intelligent calibration in comparison with laboratory calibration is its lower precision which can be improved by decreasing of the learning rate. The lower learning rate will however require longer learning time. The learning rate of the bias $B_i$ has to be much higher (100-times) than the learning rate of the gain $G_i$, otherwise the calibration stability might be corrupted and the overall calibration parameters would diverge.

## VI. CONCLUSION

In this paper we have proposed the improved sensor fusion algorithm. As an explanatory example we have used the fusion between the 3-axial gyroscope (measuring angular velocity), 3-axial accelerometer (measuring acceleration of the local system including gravity) and 3-axial magnetometer (measuring Earth's magnetic field induction) into estimation of attitude and yaw (AHRS system). The algorithm is based on the estimation of the mean square error during run-time. Then the theoretical optimal fusion gain is computed. According to Fig. 4 and Fig. 5 the adaptive gain has comparable results with the carefully chosen fixed gain fusion, but has much better dynamic characteristics (at least 5-times shorter rise time). Additionally, the parameters of adaptive systems are easier to measure directly (e.g. noise parameters of the used sensors are usually available from their manufacturer) comparing with the difficulty of proper fixed fusion gain selection. However, output MSE should be considered only as a qualitative factor since the formula used for estimation of the mean square error is only the first order approximate. Main disadvantage of the adaptive fusion algorithm is the higher CPU load (approx. 2-times more CPU time needed for MSE estimation compared with the fixed-gain based fusion algorithm).

Second part of the paper proposes the innovative run-time calibration method based on processing of fusion deviation data. The system was designed to utilize any incremental stochastic optimization (learning) method; in this article we have deployed learning method called Adam, which is an extension of the gradient descent method [24]. Because the learning algorithm is using small learning rate, it is resistant to the occasional high-power noise contained in fusion data. This feature is supported by evaluation of the mean square error of the fusion. In the discussed case of inertial attitude measurement our calibration algorithm is most suitable if the measured movement is not continuous because in a steady state the quality of estimated Euler angles rises with time. The algorithm automatically recognizes such a state and measured data have greater impact on the calibration parameters. Since the fusion is heterogeneous (absolute sensor data are merged with derivative sensor data) the bias of the absolute sensor does not affect the calibration procedure.

Although the fusion method proposed by this manuscript is derived for this special case, authors believe it can be used in many other applications as an alternative to the extended Kalman filter. The proposed fusion method is especially suitable, if the sensor readings are processed by non-linear functions in a recursive way. The proposed fusion and calibration methods can be adjusted to other sensor fusion scenarios (e.g. combination of the GNSS system – absolute velocity and a position sensor and accelerometer – a differential velocity sensor; an impulse volume sensor and a flow sensor and many others).

## ACKNOWLEDGMENT

This work has been supported by the European Regional Development Fund and the Ministry of Education of the Slovak Republic, within the project ITMS 26220220089 "New methods of measurement of physical dynamic parameters and interactions of motor vehicles, traffic flow and road."

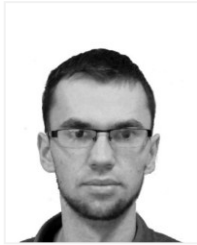
**Dušan Nemec** was born in Žilina, Slovakia in 1991. He received his MSc. degree in Automation from the University ofŽilina in 2015. He is currently pursuing the Ph.D. degree in automation at the Dep. of Control and Information Systems of the Faculty of Electrical Engineering at the University of Žilina. His research area is focused on mobile robotics and sensor systems, especially inertial navigation and flying mobile robots. In 2015 he became a laureate of the award: "Student personality of the Slovakia of the academic year 2014/2015 in the area of Electrical Engineering and Industrial Technologies".

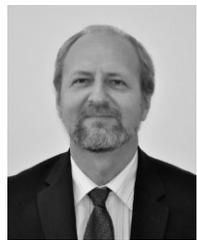
**Aleš Janota** was born in 1963 and received his MSc. degree from Technical University of Transport and Communications, Žilina, Czechoslovakia in 1981. He post-graduated from University of Zilina (UNIZA), Slovakia, in 1998 in the field of Telecommunications. From 2003 to 2009 he acted as an Assistant Professor for Information and Safety-related systems. Since 2010 he has worked as a professor in Control Engineering at the Dept. of Control and Information Systems, UNIZA. His research interests include sensors, artificial intelligence and intelligent transportation systems. From 2010 to 2014 he was a national delegate in the Domain Committee for Transport and Urban Development of the COST program.

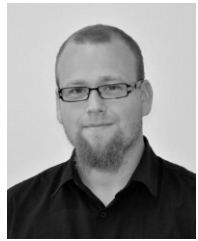
**Marián Hruboš** was born in 1987 in Martin, Slovakia. Currently works as a postdoc researcher at the Dept. of Control and Information Systems of the Faculty of Electrical Engineering at the University of Žilina (UNIZA). His research is focused on industrial automation, electrical engineering and programming, particularly on data fusion from multiple sensors for creationof 3Dmacro-textured models of the real spacesand their use in both virtual and real applications. During his Bc. and MSc. studies he was with the Institute of Competitiveness and Innovations, UNIZA. He was awarded a Certificate of Merit from the Association of Electrotechnical Industry of the Slovak Republic as a designer of the year 2014.

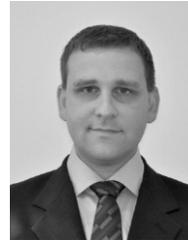
**Vojtech Šimák** was born in Žilina, Czechoslovakia, in 1980. He received the M.S. degree in information and safety-related systems (2004) and the Ph.D. degree in Control Engineering (2008) from the Faculty of Electrical Engineering, University of Žilina (UNIZA), Slovakia. During his Ph.D. study he stayed for 5 months at the Helsinki University of Technology in the Control Engineering Laboratory (2007). From 2007 to 2009, he was a researcher with the Dept. of Industrial Engineering, Faculty of Mechanical Engineering, UNIZA. Since 2009 he has worked with the Dept. of Control and Information Systems at the Faculty of Electrical Engineering, UNIZA, Slovakia. His research interests include industrial and mobile robotics and positioning systems, particularly navigation, inertial systems and computer vision.